\documentclass[10pt,twocolumn,letterpaper]{article}

\usepackage{cvpr}
\usepackage{times}
\usepackage{epsfig}
\usepackage{graphicx}
\usepackage{amsmath}
\usepackage{amssymb}
\usepackage{bm}
\usepackage{subfigure}
\usepackage{algorithm}

\usepackage[pagebackref=true,breaklinks=true,letterpaper=true,colorlinks,bookmarks=false]{hyperref}

\cvprfinalcopy % *** Uncomment this line for the final submission

\def\cvprPaperID{888} % *** Enter the CVPR Paper ID here

\ifcvprfinal\fi
\begin{document}

%%%%%%%%% TITLE
%\title{Transferring and Adapting: Learning   Semantic Representation for Unsupervised Person Re-identification}

%\title{Identity and Attribute Unsupervised Adaptation for Person Re-Identification}
%\title{Unsupervised Attribute  Adaptation for Person Re-Identification}

% domain adaptation
% Joint learning of attribute and identity
% Attribute ** 
%\title{Unsupervised Attribute Adaptation for Person Re-Identification}
%\title{Heterogeneous Learning for Unsupervised Person Re-Identification}
%\title{Unsupervised Person Re-Identification by Collaborative Learning of Attribute and Identity}
\title{Transferable Joint Attribute-Identity Deep Learning for \\ Unsupervised Person Re-Identification}

% Unsupervised Cross-Dataset Transfer Learning

%\title{Semantic Knowledge Transfer and Collaborative Learning for Person Re-identification with Attributes}

%\title{Cross Task Knowledge Transfer and Collaborative Learning for Person Re-identification with Attributes}
%\title{Learning Transformable Semantic Representation by Knowledge Transfer and Collaborative Learning for Person Re-identification}

\author{Jingya Wang$^1$ 
	\quad \quad \quad \quad Xiatian Zhu$^2$ 
	\quad \quad \quad \quad Shaogang Gong$^1$ 
	\quad \quad \quad \quad Wei Li$^1$ \\
Queen Mary University of London$^1$ \quad \quad \quad \quad %\\
% London E1 4NS, United Kingdom 
Vision Semantics Ltd.$^2$\\
{\tt\small \{jingya.wang, s.gong, w.li\}@qmul.ac.uk} 
\quad \quad 
\tt\small eddy@visionsemantics.com
}

\maketitle
%\thispagestyle{empty}

%%%%%%%%% ABSTRACT
\begin{abstract}

Most existing person re-identification (re-id) methods require supervised model learning 
from a separate large set of pairwise labelled training data for every single camera pair.
This significantly limits their scalability and usability in real-world 
large scale deployments with the need for performing re-id across many camera views. 
To address this scalability problem, we develop a novel 
deep learning method for transferring the labelled information of
an existing dataset to a new unseen (unlabelled) target domain for
person re-id without any supervised learning in the target domain.
Specifically, we introduce an Transferable Joint Attribute-Identity Deep Learning
(TJ-AIDL) for simultaneously learning an attribute-semantic and identity-discriminative
feature representation space transferrable to 
any new (unseen) target domain for re-id tasks 
without the need for collecting new labelled training data from the target
domain (i.e. unsupervised learning in the target domain).
Extensive comparative evaluations validate the superiority of this new TJ-AIDL model 
for unsupervised person re-id over a wide range of state-of-the-art methods on four challenging benchmarks including
VIPeR, PRID, Market-1501, and DukeMTMC-ReID. 
	
%	
%Person Re-Identification is a challenging task in computer vision, especially no training data available for target camera views. In this paper, we propose a deep learning based methods by transferring high-level identification knowledge into middle level semantic representation, which has strong cross domain generalization ability. Moreover, when adapt to new target domain without supervision, single task classifier is easily drift to previous domain knowledge(domain drift problem), we present a deep collaborative learning (DCL) strategy: two students, one student with identification prior knowledge, the other with attribute prior knowledge, collaborative learning in semantic space on new target domain and adjust itself to make finial agreement. Experiment results show that our methods achieve state of art result on Person Re-Identification in unsupervised setting.

\end{abstract}

%%%%%%%%% BODY TEXT
\section{Introduction}

Person re-identification (re-id) aims at matching people across non-overlapping camera views
distributed at distinct locations.
Most existing re-id studies follow the {\em supervised} learning paradigm
such as 
optimising pairwise matching distance metrics \cite{KISSME_CVPR12,xiong2014person,PRDC,wang2014person,zhang2016learning,wang2016human,wang2016pami,chen2017person}
or deep learning methods
\cite{li2014deepreid,wangjoint,xiao2016learning,
subramaniam2016deep,chen2016multi,li2017person,
li2018HAC_reid,chang2018MLFN_reid}.
They assume the availability of a large number of {\em manually}
labelled matching pairs for each pair of 
camera views for learning a feature representation or a matching
distance function optimised for that camera pair.
However, this leads to a poor scalability in 
practical re-id deployments, because
such scale manual labelling 
is not only prohibitively expensive to collect in the real-world as there are
a quadratic number of camera pairs, but also implausible in many cases,
e.g. there may not exist sufficient
training people reappearing in every pair of camera views.
This scalability limitation severely 
reduces the usability of existing supervised re-id methods.

One generic solution to large scale re-id in real-world deployment is
designing {\em unsupervised} models.
While a few unsupervised methods
have been developed
\cite{farenzena2010person,cheng2011custom,kodirov2015dictionary,kodirov2016person,lisanti2015person,wang2014unsupervised,zhao2017person},
they typically offer weaker re-id performances when compared to the supervised counterparts.
This makes them less useful in practice.
One main reason is that without labelled data across views,
unsupervised methods lack the necessary knowledge on how visual
appearance of identical objects changes cross-views
due to different view angles, background and illumination. 
%labelled matching pairs across camera views, existing un- supervised models are unable to learn what makes a person recognisable under severe appearance changes.
%
Another solution is to exploit simultaneously 
{\bf (1)} unlabelled data from a target domain 
%(in the re-id context, thousands of people may travel through each camera %view everyday in a busy public space) 
and 
{\bf (2)} existing labelled datasets from some training source domains. 
Specifically, the idea is to learn a feature representation that contains some
view-invariant information about people appearance learned from labelled source data,
transfer and adapt it to a target domain by using only unlabelled
target data for re-id matching in the target domain. 
As the target dataset has no label, 
this is regarded as an {\em unsupervised learning} problem. % \cite{peng2016unsupervised,yu2017cross}.

There are a few studies on exploiting unlabelled target data 
for unsupervised re-id modelling using either identity or attribute label, or both
from source datasets
\cite{peng2016unsupervised,yu2017cross,su2016deep}.
However, they generally offer weaker re-id performance 
due to either domain sensitive hand-crafted features
or a lack of an effective knowledge transfer learning algorithm
between attribute and identity discriminative features.
It is very challenging to address
this cross-domain and multi-task (between attribute and identity)
transfer learning problem in a principled way due to three co-occurring uncertainties:
(1) Source and target domains have unknown camera viewing conditions;
(2) The identity/class population between source and target
domains are non-overlapping therefore presents a more challenging
open-set recognition problem, as compared to the closed-set assumption made
by most existing transfer learning models \cite{pan2010survey};
(3) Joint exploitation of attribute and identity labels
suffers from the heteroscedasticity
(a mixture of different knowledge granularity and characteristics) 
learning problem \cite{duin2004linear}.

In this work, we consider unsupervised person re-id
by sharing the source domain knowledge through attributes learned from
labelled source data and transfering such knowledge to unlabelled target data by a
joint attribute-identity transfer learning across domains.
We make three {\bf contributions}:
{\bf (I)} 
We formulate a novel idea of {\em heterogeneous multi-task joint deep learning} 
of attribute and identity discrimination for unsupervised person re-id.
%in order to address the under-studied deployment scalability issue. 
%
%
To our best knowledge, this is the first attempt at {\em joint deep learning
of auxiliary attribute and identity labels} 
for solving the unsupervised person re-id problem cross-domains. 
{\bf (II)}
We propose a {\em Transferable Joint Attribute-Identity Deep Learning} (TJ-AIDL)
to simultaneously learn {\em global} identity 
and {\em local} attribute information from labelled source domain person images
through an Identity Inferred Attribute (IIA) space
for maximising the joint learning effectiveness
between identity and attribute.
This IIA is designed specially to address the
notorious heteroscedasticity challenge from which
the common space multi-task joint learning often suffers.
Importantly, the IIA interacts concurrently with both the attribute and identity learning tasks
inter-dependently without breaking 
the end-to-end model learning process.
{\bf (III)}
We introduce an attribute consistency scheme
for performing TJ-AIDL model unsupervised adaptation on the unlabelled target data
to further enhance its discriminative compatibility towards
each target domain re-id task at hand.
Extensive evaluations demonstrate the superiority of the
proposed TJ-AIDL model over a wide range of state-of-the-art
re-id models on four challenging benchmarks
VIPeR~\cite{ELF_ECCV08},
PRID \cite{hirzer2011person},
Market-1501~\cite{zheng2015scalable},
and DukeMTMC-ReID \cite{zheng2017unlabeled}.

%-------------------------------------------------------------------------
%-------------------------------------------------------------------------

\section{Related work}

\noindent {\bf Person Re-ID}
Most existing re-id models are based on {\em supervised} learning
for every camera pair
on a separate set of labelled training data
\cite{KISSME_CVPR12,xiong2014person,PRDC,wang2014person,
zhang2016learning,wang2016human,wang2016pami,chen2017person,li2014deepreid,
wangjoint,xiao2016learning,subramaniam2016deep,
chen2016multi,li2017person,chen2017personICCVWS,li2018HAC_reid,chang2018MLFN_reid}.
They suffer from
poor scalability in 
realistic re-id deployments
where no such a large training set is available
for each single camera pair.
To solve this scalability issue,
unsupervised methods based on hand-crafted features
\cite{farenzena2010person,cheng2011custom,kodirov2015dictionary,kodirov2016person,lisanti2015person,wang2014unsupervised,zhao2017person,wang2014unsupervised} can be chosen for deployment.
However, they usually yield much weaker performance
than supervised models therefore practically not very useful.
While a balance between scalability and matching accuracy
can be achieved by semi-supervised learning,
existing methods \cite{liu2014semi,wang2016towards} still demand a fairly large set of pairwise labels 
which is again not scalable.

Recently, unsupervised re-id by cross-domain transfer learning has been developed 
to exploit labelled data from source datasets by extracting
transferable identity-discriminative information to an unlabelled target dataset \cite{peng2016unsupervised,yu2017cross,su2016deep}.
However, these methods have a few limitations that restrict their generalisation:
(1) Relying on hand-crafted features 
without the deep learning capability of automatically 
learning stronger representations from training data
\cite{peng2016unsupervised};
(2) Using a pre-learned deep model on labelled source data
but lacking an effective domain adaptation mechanism \cite{yu2017cross};
(3) Independently exploiting identity and attribute label supervision
in model learning therefore ignoring their interaction and compatibility
\cite{su2016deep}.
Data synthesis \cite{an2017multi,deng2018image} has also been
proposed as a solution for addressing limited data, although it
suffers from undesirable person appearance distortion and restricted source selection. 
%\textcolor{red}{More recently, synthetic data generation  method \cite{an2017multi} and GAN based domain adaptation method \cite{deng2018image} were introduced, however, still limit in global identity discrimination learning and ignored transferable ability for local semantic representation.}
%
The proposed TJ-AIDL method addresses these limitations of existing methods
in a unified deep joint learning model.
Moreover, our method goes beyond the common multi-task joint learning
design by introducing a more transferable mechanism for
discriminatively optimising both attribute and identity learning in a shared end-to-end process.
Our experiments show that the proposed method significantly 
outperforms existing models even by using less supervision in the source domain.

\vspace{0.1cm}
\noindent {\bf Attribute for Re-ID}
%\textcolor{red}{Several works have been done for pedestrian
%attributes recognition \cite{wang2017attribute,liu2017hydraplus} and these}
Visual semantic attributes \cite{wang2017attribute,liu2017hydraplus} have been exploited as a mid-level feature representation for cross-view re-id 
\cite{layne2012person,layne2014attributes,layne2014re,su2016deep,su2015multi,peng2016unsupervised}. 
However, such semantic coefficient representations are less
powerfull for identity discrimination than conventional feature vectors. 
The reasons are: 
(1) Attribute coefficient representations are usually of low dimensions 
(tens vs. thousands for typical low-level feature representations)
\cite{zheng2015scalable,liao2015person,ELF_ECCV08,PRDC}; 
(2) Consistently predicting individually all the attributes is a
difficult task when the labelled training data is sparse and person
images have low quality as mostly in person re-id datasets, that is,
inter-attribute discrimination can be weak on typical person re-id images.
%
%However, this does not mean attributes cannot be mined in a deeper way.
%
To overcome these problems, we explore attributes in our TJ-AIDL model
by introducing a mechanism to extract identity discriminative
attribute information through co-learning both attribute and identity labelled data jointly. 
Moreover, we uniquely employ the attribute space for unsupervised domain adaptation.% from the view consistency perspective.

\begin{figure*}%[th!]
	\centering
	\includegraphics[width=1\linewidth]{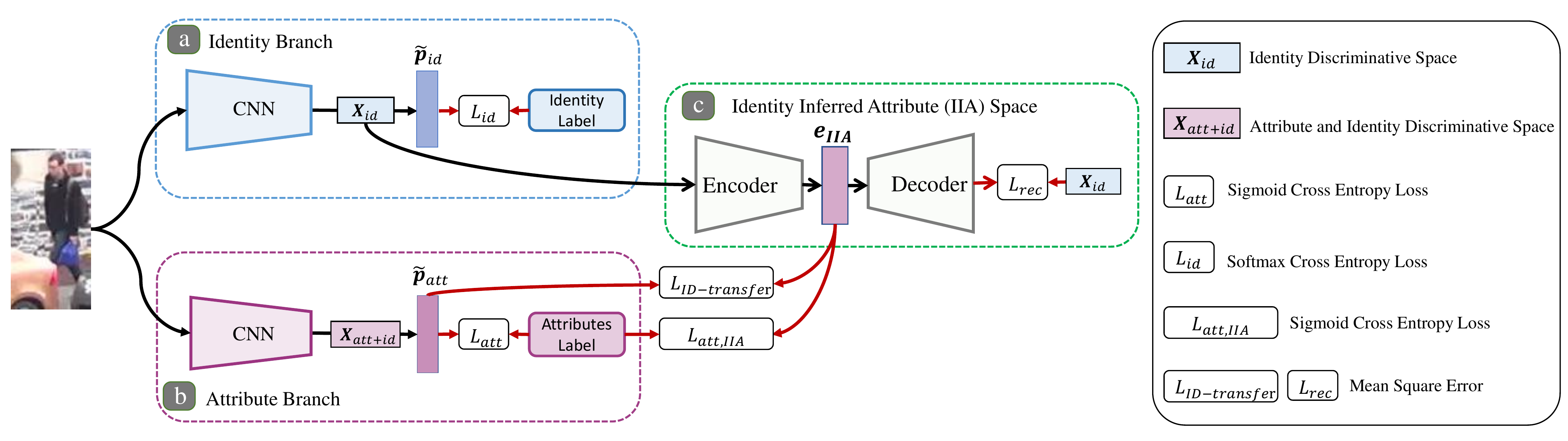}
	% \vskip -0.4cm
	\caption{%\footnotesize
	An overview of the proposed Transferable Joint Attribute-Identity Deep Learning (TJ-AIDL).
}
	\label{fig:pipeline}
	\vspace{-.0cm}
\end{figure*}

%-------------------------------------------------------------------------
\section{A Joint Attribute-Identity Space}

\noindent {\bf Problem Definition}
For person re-id by attribute (semantic) based unsupervised domain adaptation,
we have a {\em supervised} source dataset (domain) 
$\{(\bm{I}_i^s, y_i^s, \bm{a}_i^s)\}_{i=1}^{N^s}$ consisting of $N^s$ person 
bounding box images $\bm{I}^s$, 
the corresponding {\em identity} $y^s \in \mathcal{Y}=\{1,\cdots,N^s_\text{id}\}$ 
(i.e. a total $N_\text{id}^s$ different persons), and 
identity-level {\em binary attribute} $\bm{a}_s \in \mathcal{R}^{m\times1}$ 
(i.e. a total $m$ different attributes)
labels.
We also assume a set $\{\bm{I}_i^t\}_{i=1}^{N^t}$ of $N^t$ unlabelled target training data,
which can be used for model domain adaptation.
The {\bf objective} is to develop an unsupervised domain adaptation approach 
to learning the optimal feature representation
by transferring the supervised identity and attribute knowledge of the source domain 
to person re-id in a target domain with only {\em unlabelled} data of
entirely a different pool of identity classes.
Note that: 
{\bf (1)} Unlike identity label, 
attribute detection is a {\em multi-label} recognition problem since
the $m$ attribute categories co-exist in every single person
image.
{\bf (2)}
These two types of label supervision
lie at quite different levels: Most attributes are {\em localised} to image
regions, even though the location information is not
provided in the annotation; While
person identity labels are at the {\em holistic} image-level.
It is a non-trivial learning task since this is not only a multi-label learning problem -- joint
learning of mutually correlated attribute labels, 
but also a
{\em heterogeneous multi-task joint learning} problem -- inter-dependently
learning a person re-id representation space by joint holistic identity and 
local attribute supervision.

In this work, we present a novel {\em Transferable Joint Attribute-Identity Deep Learning} (TJ-AIDL) approach to 
establishing an identity-discriminative and attribute-sensitive (i.e. dually-semantic) feature representation
space optimal for person re-id on the labelled target domain without any identity and attribute labels provided.
We avoid simply combining re-id and attribute feature vectors in deep model design to gain their 
complementary advantages, which may suffer from the heteroscedasticity 
% (a mixture of different knowledge granularity and characteristics)
problem \cite{duin2004linear}
and finally results in sub-optimal results.
Instead, we assign them into two separate branches for simultaneously learning
individual discriminative features subject to the corresponding 
label supervision concurrently.
Importantly, we design a progressive knowledge fusion mechanism by 
introducing an {\em  Identity Inferred Attribute} (IIA) regularisation space
%acting as a model learning regularisation, 
for more smoothly
% more discriminatively 
transferring the global identity information into 
the local attribute feature representation space.
It is also the proposed IIA space that provides an opportunity
that allows for adapting the learned model to the target domain
where no identity and attribute labels are available. 
%
%During unsupervised target domain adaptation, 
%the proposed IIA provides a means for 
%
% which results in a dually-semantic 
%feature representation optimal for 
%
As such, the proposed TJ-AIDL largely addresses the joint learning challenges of
heterogeneous identity and attribute label information sources in a shared representational space
in a more challenging cross-domain context.
%best individual attribute prediction given both local and
%holistic cross-domain annotations.

%First, we define the problem setting and corresponding notation. 
%=============
%
%Assume on source dataset, $D_{s}= \{X_{s},D_{s},A_{s}\}$, with number $n$ input images $X_{s}= [x_{s1},...x_{sn}]$and corresponding ID labels $D_{s}= [d_{s1},...d_{sn}]$ and attribute labels $A_{s}= [a_{s1},...a_{sn}]$, where $a_{si} = \{a^{_{si}^{1}},...a^{_{si}^{m}}\} $ and $m$ is the number of attributes. Our goal is given a new target dataset $D_{t}$ which contains completely different set of identities/classes without any label, learning the optimal representation/matching function for the target dataset using the knowledge transferred from the labbled source dataset.

\subsection{Transferable Joint Deep Learning}

\noindent {\bf Model Overview} 
We consider a multi-branch network architecture 
for our heterogeneously supervised multi-task learning.
The rational of this multi-branch composition 
is to maintain a sufficient independence of each supervision learning tasks
for avoiding their potentially negative mutual influence due to their {semantic 
discrepancy}. 
%
%\subsection{Cross-Task Intermediate Semantic Transfer}
An overview of the proposed Transferable Joint  Attribute-Identity Deep Learning (TJ-AIDL) method
is depicted in Fig. \ref{fig:pipeline}. 
The TJ-AIDL contains two branches:
{(1)} {\em Identity Branch}: which aims to extract the re-id sensitive information from
the available identity class labels in the source domain (Figure \ref{fig:pipeline}(a)).
{(2)} {\em Attribute Branch}:
which aims to extract the semantic knowledge from the attribute labels 
(also from the source domain) (Figure \ref{fig:pipeline}(b)).
To establish a channel for knowledge fusion,
we introduce the {\em  Identity Inferred Attribute } (IIA) space (Figure \ref{fig:pipeline}(c))
designed for transferring the re-id discriminative information from the Identity Branch 
to the Attribute Branch where two-source information is synergistically 
integrated in a {\em smoother} manner.
That is, once the TJ-AIDL is trained, 
the feature representations extracted from the Attribute Branch can be 
directly exploited for re-id deployment.

For unsupervised person re-id by cross-domain knowledge transfer and target data adaptation, 
we conduct the model training of our proposed 
TJ-AIDL in two steps:
{\bf (I)} {\em Attribute-Identity Transferable Joint Learning}:
This is supervised by the source labelled training data;
{\bf (II)} {\em Unsupervised Target Domain Adaptation}:
This is performed on the target unlabelled training data.
We describe more details for each component of our TJ-AIDL
in two training steps.

%It has three sub-network branches,:(1)ID branch that learning the discriminative visual representation (2)Intermediate transfer branch that transfer identification knowledge into intermediate semantic space (3) attribute branch that learning the semantic representation.

%It has three sub-network branches,:(1)ID branch that learning the discriminative visual representation (2)Intermediate transfer branch that transfer identification knowledge into intermediate semantic space (3) attribute branch that learning the semantic representation.

%\subsubsection{Learning discriminative visual representation }

\subsubsection{Attribute-Identity Transferable Joint Learning}
\label{sec:method_TJL}

\noindent{\bf Identity and Attribute Branches}
For building an efficient yet strong deep re-id model,
we choose the lightweight MobileNet as the CNN architecture\footnote{
	This selection is independent of our model design
	and others can be readily applied, e.g.
%	AlexNet \cite{krizhevsky2012imagenet},
	ResNet \cite{he2016deep}, Inception \cite{szegedy2016rethinking} and VggNet \cite{simonyan2014very}.}
for both identity and attribute branches.
%
%We construct the ID branches using Mobilenet architecture
%as our base-net, other structure can be readily applied.
For training the identity branch (Fig. \ref{fig:pipeline}(a)), we use the softmax Cross Entropy 
loss function defined as:
%\begin{equation}
%p_{i}^{n}(x_{i})=\frac{exp(z_{i}^{n}))}{\Sigma_{k=1}^{n}exp(z_{i}^{n}))}
%\label{key}
%\end{equation}
%where $z_{i}$ refers to is the output of the “softmax” layer , and the training loss is computed as:
\begin{equation}
L_\text{id}=-\frac{1}{n_\text{bs}}\sum_{i=1}^{n_\text{bs}}
%\log \frac{\exp(z_{i}^{n}))}{\Sigma_{k=1}^{n}\exp(z_{i}^{n}))}
\log \Big( p_\text{id}(\bm{I}_{i}^s, y_i^s) \Big)
\label{eq:CE_softmax}
\end{equation}
where $p_\text{id}(\bm{I}_{i}^s, y_i^s)$ specifies the predicted probability on the groundtruth 
class $y_i^s$ of %the training image 
$\bm{I}_{i}^s$,
and $n_\text{bs}$ denotes the batch size.

Given that the attribute branch (Fig. \ref{fig:pipeline}(b)) is a multi-label classification learning task,
we instead use the Sigmoid Cross Entropy loss function to generate the training signal
by considering all $m$ attribute classes:
\begin{align}\label{eq:CE_sigmoid}
\small
L_\text{att}=-\frac{1}{n_\text{bs}}\sum_{i=1}^{n_\text{bs}}
\sum_{j=1}^{m}
\Big( 
a_{i,j} \log\big( p_\text{att}(\bm{I}_i,j) \big) + \\ \nonumber
(1-a_{i,j}) \log \big( 1 - p_\text{att}(\bm{I}_i,j) \big)
\Big)
\end{align}
where $a_{i,j}$ and $p_\text{att}(\bm{I}_i,j)$ 
define the groundtruth label
and the predicted classification probability on the $j$-th attribute class
of the training image $\bm{I}_i$,
i.e. $\bm{a}_i = [a_{i,1},\cdots,a_{i,m}]$
and $\bm{p}_{\text{att},i} = [p_\text{att}(\bm{I}_i,1),\cdots,p_\text{att}(\bm{I}_i,m)]$.

By {\em independently} training the two branches using the above designs,
we only allows for optimising their respective features without
exploiting their complementary effect for maximising
the compatibility.
A common approach is to build a multi-task joint learning network
which {\em directly} subjects a {\em shared} feature representation to both 
identity loss (Eq. \eqref{eq:CE_softmax})
and attribute loss (Eq. \eqref{eq:CE_sigmoid}) 
concurrently in model training.
Instead, we present an alternative progressive scheme for 
more effective multi-source knowledge fusion 
as described below (see evaluations
in Sec. \ref{sec:exp_fusion_method}).

%to optimise the identity branch
%(a single label classification task)
%and the Sigmoid Cross Entropy loss function
%to learn the attribute branch
%(a multi-label classification task).
% 
%discrimination given training labels of multiple person
%classes extracted from pair-wise labelled re-id dataset. 

%\begin{equation}
%p_{i}^{n}(x_{i})=\frac{exp(z_{i}^{n}))}{\Sigma_{k=1}^{n}exp(z_{i}^{n}))}\\
%\end{equation}
%
%where $z_{i}$ refers to is the output of the “softmax” layer , and the training loss is computed as:
%\begin{equation}
%L_{id}=-\frac{1}{n}\sum_{i=1}^{n}log(p_{i}^{n}(x_{i}))
%\end{equation}

%Finally, Given the image $I$, Mobilenet generate 1024*1 feature vector, denote as $v_{i}$.

%Note: The ID branch network is independent of our model design, we can use any off-the shelf re-id model to extract the feature.

\vspace{0.1cm}
\noindent{\bf  Identity Inferred Attribute Space}
We introduce an intermediate 
Identity Inferred Attribute (IIA) Space 
for achieving the knowledge fusion learning on
attribute and identity labels in a softer manner
(Fig. \ref{fig:pipeline}(c)).
The IIA space is jointly learned with the two branches 
while being exploited to perform information transfer and fusion from the identity branch
to the attribute branch simultaneously.
This scheme allows for both consistent and cumulative knowledge fusion in the whole training course.
%
%At the same time, the features are given space to adjust to the most recent environment as only their projection into a low dimension submanifold is controlled. The proposed system is evaluated on image classification tasks and shows a reduction of forgetting over the state-of- the-art.

More specifically, we build the IIA space in the encoder-decoder (auto-encoder) framework
due to that: (1) It has a strong capability of capturing the most important information 
of a given target task (represented by the input data)
via a concise feature vector representation;
(2) More importantly, such a concise feature representation
facilitates the inter-task information transfer whilst still
preserving sufficient updating freedom space to every individual learning task
\cite{bourlard1988auto,rannen2017encoder}.
We call this sub-model {\em IIA encoder-decoder}.
In our context, we want to extract and compress essential identity information
into the IIA space for facilitating fusion.
We therefore exploit the identity features (Fig. \ref{fig:pipeline}(a))
as the input of IIA encoder and also
the groundtruth of IIA decoder 
(i.e. reconstruction unsupervised learning).
Once the input is given, this model itself can be
learned based on the reconstruction loss (Mean Square Error (MSE)): % in an unsupervised manner:
\begin{equation}
L_\text{rec}  = \| \bm{x}_\text{id} - f_\text{IIA}(\bm{x}_\text{id}) \|^2 
%E((z_{p} - z_{p^{'}})^{2})
\label{eqn:}
\end{equation}
where $\bm{x}_\text{id}$ represents the identity feature of a training image
and $f_\text{IIA}()$ the mapping function of IIA encoder-decoder.
By this unsupervised learning manner, 
we are able to obtain a latent feature embedding $\bm{e}_\text{IIA}$ 
with important identity information encoded. 
To transfer the identity information across branches, 
we need a corresponding low dimensional matchable space in the attribute counterpart,
which however is not available.

To address the above problem, we propose to align the IIA embedding $\bm{e}_\text{IIA}$ 
with the prediction distribution over all $m$ attribute classes,
in spirit of knowledge distillation \cite{hinton2015distilling}.
As such, we naturally set $m$ as the dimension of $\bm{e}_\text{IIA}$ for easing alignment and 
cross-branch knowledge transfer without the need for an additional transformation.

More formally, we conduct the identity knowledge transfer via imposing an
MSE based identity transfer loss:
\begin{equation}
L_\text{ID-transfer}  = \| \bm{e}_\text{IIA} - \bm{\tilde{{p}}}_\text{att} \|^2
\label{eq:ID_transfer}
\end{equation}
where $\bm{\tilde{{p}}}_\text{att}$ is logits from the attribute branch. 
Considering that the $\bm{e}_\text{IIA}$ is derived in an unsupervised manner
which may be over further way from the attribute prediction counterpart and hence 
giving a harder alignment task,
we add similarly a sigmoid Cross Entropy loss to the learning of $\bm{e}_\text{IIA}$
by exploiting it as a pseudo attribute prediction, as
%\begin{align} \small
%L_\text{attr, IIA}  = -
%%\frac{1}{n_\text{bs}}\sum_{i=1}^{n_\text{bs}}
%%\sum_{j=1}^{m}
%%\Big( 
%a_{i,j} \log\big( e_{i,j} \big) + % \\ \nonumber
%(1-a_{i,j}) \log \big( 1 - e_{i,j} \big)
%%\Big)
%\label{eq:pseudo_CE}
%\end{align}

\begin{align} \small
%\color{red}
L_\text{attr, IIA}=-\frac{1}{n_\text{bs}}\sum_{i=1}^{n_\text{bs}}
\sum_{j=1}^{m}
\Big( 
a_{i,j}\log\big( p_\text{IIA}(\bm{I}_i,j) \big) + \\ \nonumber
(1-a_{i,j}) \log \big( 1 - p_\text{IIA}(\bm{I}_i,j) \big)
\Big)
\end{align}
{where $ p_\text{IIA}(\bm{I}_i,j)$ is the probability predicted based on $\bm{e}_\text{IIA}$ by the sigmoid function.}
%where $\bm{e}_{\text{IIA}, i} = [e_{i,1}, \cdots, e_{i,m}]$
%is the IIA embedding of the training image $\bm{I}_i$.
Finally, we formulate the overall IIA loss function by incorporating the above components
by weighted summation as:
\begin{equation}
L_\text{IIA} = L_\text{attr, IIA} + \lambda_{1}L_\text{rec} + \lambda_{2}L_\text{ID-transfer}
\label{eq:IIA_loss}
\end{equation}
where $\lambda_1$ and $\lambda_2$ are scale normalisation parameters to %empirically fix in our evaluations.
ensure all three loss quantities are of a similar scale in value.
%\textcolor{red}{keep the losses in same scale and empirically fix in our experiments.}

\vspace{0.1cm}
\noindent {\em Impact of IIA on Identity and Attribute Branches}
The introduction of IIA imposes different influence on the two branches in model training.
Since IIA is established on the identity features,
no change is imposed into the learning of this branch.
For the attribute branch, however, an additional learning constraint 
is created for identity knowledge transfer.
We therefore reformulate its supervised learning loss function
by incorporating Eq. \eqref{eq:ID_transfer} as:
\begin{align}\label{eq:att_loss_IIA}
\small
%\color{red}
L_\text{att-total}=L_\text{att} + 
\lambda _2 L_{\text{ID-transfer}}
\end{align}

\vspace{0.1cm}
\noindent {\bf Remarks}
The IIA component aims at creating an interactive learning mechanism
between the identity and attribute branches in a more transferable way.
This significantly differs from the straightforward joint learning approach
which suffers from the underlying multi-source information incompatible problem.
We summarise the main information flow in model joint training:
(1) The identity branch learns to extract identity discriminative information;
(2) The IIA component then transfers the identity information to the attribute branch;
(3) The attribute branch learns to extract attribute discriminative knowledge whilst
simultaneously incorporating/fusing identity sensitive information.  
%when the identity information is transferred into the attribute branch via the IIA component,
%
However, the TJ-AIDL model learned on the labelled source data is still not optimal 
for re-id in a typically {\em unlabelled} target domain due to the inevitable presence of 
domain shift in real-world deployment scenarios.
This leads to the necessity of model unsupervised domain adaptation, as detailed below.

\subsubsection{Unsupervised Target Domain Adaptation}
%\subsection{Cross Domain Adaptation}
\label{sec:method_adaptation}

\begin{figure}%[th!]
	\centering
	\subfigure[Source data supervised learning of TJ-AIDL by attribute consistency]{
%		\rule{4cm}{3cm}
		\includegraphics[width=1\linewidth]{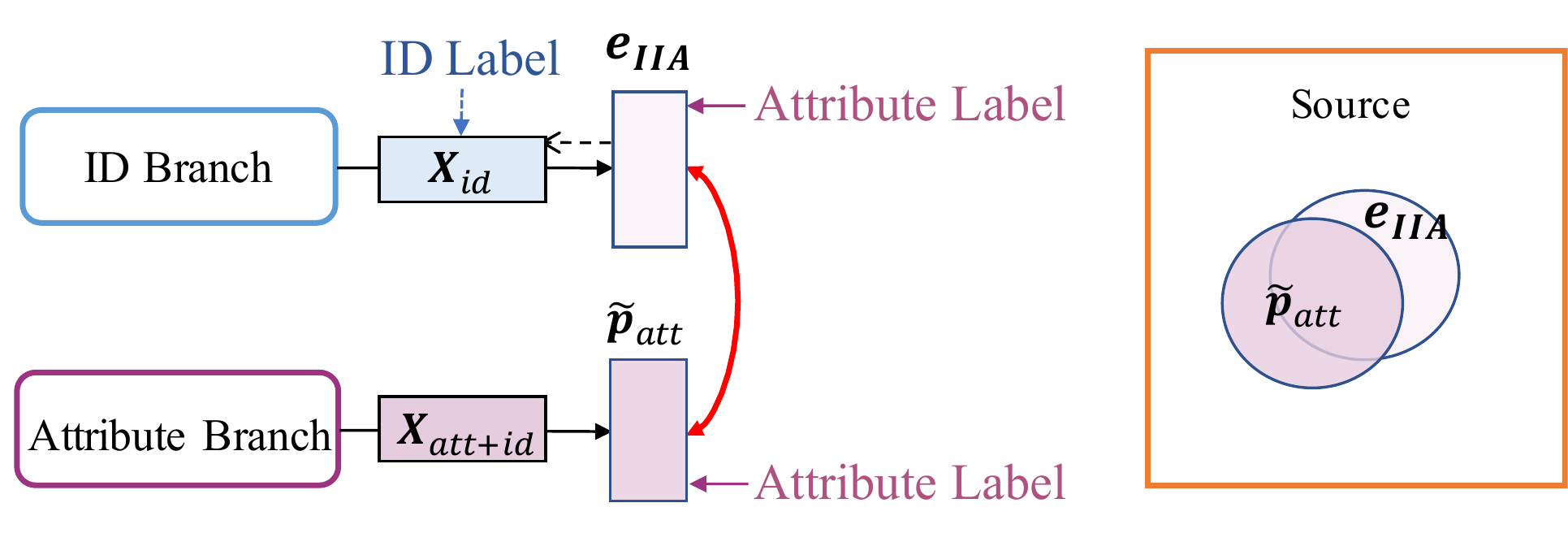}
		\label{fig:stage1}
	}
%\vskip -0.1cm
	\subfigure[{Taget domain adaptation} of TJ-AIDL by attribute consistency
%	 in the target domain.
	 ]{
		%		\rule{4cm}{3cm
		\includegraphics[width=1\linewidth]{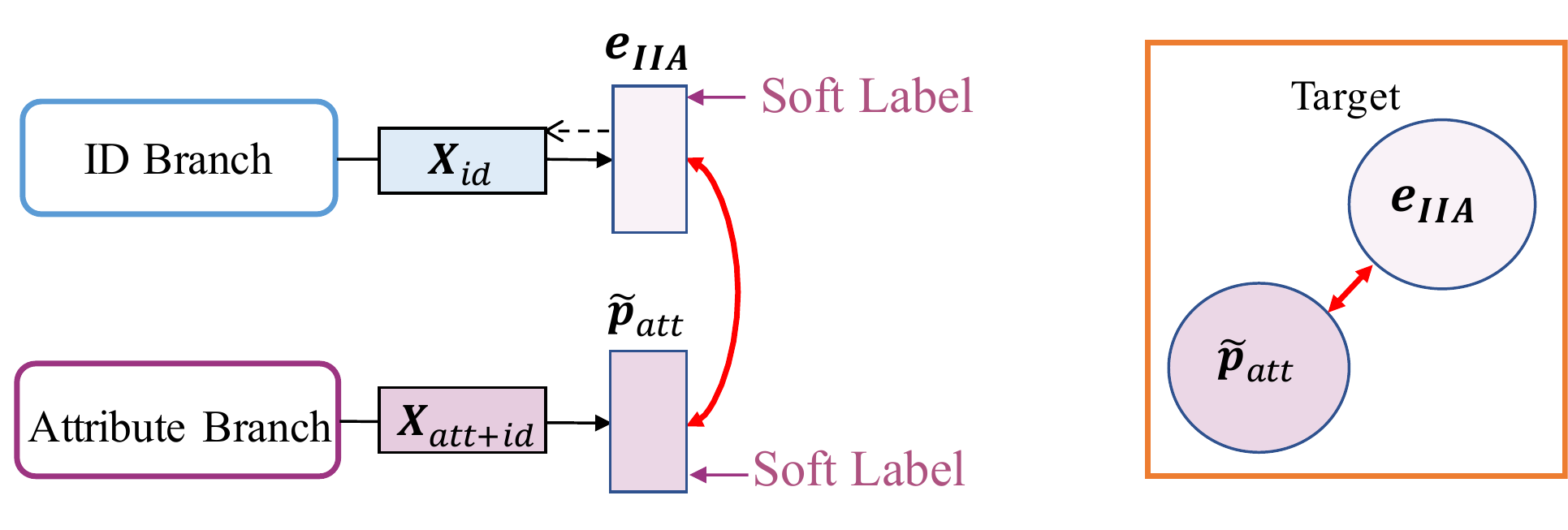}
		\label{fig:stage2}
	}
	% \vskip -0.4cm
	\caption{%\footnotesize
	An illustration of the attribute consistency maximisation idea
	for unsupervised target domain adaptation. 
	Given a TJ-AIDL model trained on the source domain,
	it has more attribute consistency (a) on the source domain, 
	(b) but less on the unseen target domain. See more details in
        the main text.
}
	\label{fig:stage}
	\vspace{-.0cm}
\end{figure}

We want to adapt a learned TJ-AIDL model to fit
the unlabelled target domain data.
To that end, we exploit the attribute consistency principle
by treating the prediction of attribute branch and 
the embedding of IIA component 
as different attribute perspectives from different domains. 
This idea is based on the observation that,
a well fitted TJ-AIDL model is supposed to have small discrepancy 
between the two different attribute perspectives, for example, the one trained on the source domain 
(Fig. \ref{fig:stage}(a)). 
In other words, their consistency degree suggests how well the model fits a given domain.
This also partially shares the spirit of the cyclic consistency mechanism \cite{sener2016learning}.

Specifically, our objective is to adapt the attribute branch since it is used in re-id deployment.
Hence, we can ignore the updating of the identity branch.
We design the following adaptation algorithm:
{\bf (1)} We deploy the TJ-AIDL model learned on the source domain 
on unlabelled target person images to obtain the attribute prediction 
$\bm{p}_{\text{att},t}$ from the attribute branch.
{\bf (2)} We then utilise the {soft label} $\bm{p}_{\text{att},t}$ as the pseudo groundtruth to
update both the attribute branch and IIA component
for reducing attribute discrepancy between domains (Fig. \ref{fig:stage}(b)). 
Intuitively, this soft attribute label is needed since
we need to prevent the model drifting overly by maintaining the 
most attribute discriminative power obtained from the source domain.
{\bf (3)} We adapt the model on the target training data until convergence.

%%%Update  $L_{att}$ by soft label(todo)
%%
%%\begin{equation}
%%L_{s^{'}} = L_{att\_soft}+w_{1}*L_{reconstruct}+w_{2}*L_{consist}
%%\end{equation}
%%\begin{equation}
%%L_{s} = L_{att\_soft}+w_{2}*L_{consist}
%%\end{equation}
%
%Collaborative learning is a situation in which two or more people learn or attempt to learn something together.[1] Unlike individual learning, people engaged in collaborative learning capitalize on one another's resources and skills.
%Experts from different field interact with each other and finally make agreement ,found this to be beneficial in helping students learn effectively and efficiently than if the students were to learn independently. Some positive results from collaborative learning activities are students are able to learn more material by engaging with one another and making sure everyone understands.

%\begin{figure}%[th!]
%	\centering
%	\includegraphics[width=1\linewidth]{image/Collaborative Learning.pdf}
%	% \vskip -0.4cm
%	\caption{%\footnotesize
%	Collaborative Learning	}
%	\label{fig:problem}
%	%\vspace{.5cm}
%\end{figure}

\subsection{Model Optimisation and Deployment}

\noindent {\bf Optimisation}
Our TJ-AIDL model can be trained using the standard Stochastic Gradient Descent algorithm
in end-to-end manner. 
%Our implementation is carried out in Tensorflow \cite{abadi2016tensorflow} which supports automatic differentiation through the gradient updates. 
We summarise the training process in Alg. \ref{Algorithm}.

\vspace{0.1cm}
\noindent {\bf Deployment}
Given a TJ-AIDL model trained on a labelled source domain and adapted on the unlabelled
target domain,
we obtain a 1,024-D deep feature representation from the attribute branch (Fig. \ref{fig:pipeline}(b)).
This feature vector is not only attribute semantic but also identity discriminative.
Hence, we deploy this 1,024-D deep feature 
for person re-id deployment by the $L_2$ distance
in the target domain.

%(1)ID branch $p$ (can replace by off-the-shelf id feature)
%
%(2)intermediate branch $s^{'}$
%
%(2)semantic branch $s$

\begin{algorithm}[h]
	\small
	\caption{\small Learning the TJ-AIDL model.} \label{Algorithm}
	\parbox{3.25in}{ 
		\textbf{Input:} $N^s$ labelled source $\{(\bm{I}_i^s, y_i^s, \bm{a}_i^s)\}_{i=1}^{N^s}$ and 
		$N^t$ unlabelled target $\{\bm{I}_i^t\}_{i=1}^{N^t}$ training data;
		\\ [0.05cm]
		\textbf{Output:} TJ-AIDL re-id model; %$\bm{\Theta}$
		\\[0.05cm]
		\textbf{Step I: Transferable Joint Learning (Sec. \ref{sec:method_TJL})}  
		\\[0.05cm]
		\textbf{for} $t=1$ \textbf{to}  \textsl{max-iteration} \textbf{do} \\ 
		\hphantom{~~~~~~}
		Sampling a batch of labelled source data; \\
		\hphantom{~~~~~~} 
		Identity branch evaluation (samples feed-forward);\\
		\hphantom{~~~~~~} 
		Attribute branch evaluation;\\
		\hphantom{~~~~~~} 
		Updating the identity branch (Eq. \eqref{eq:CE_softmax});\\
		\hphantom{~~~~~~} 
		Updating the IIA encoder-decoder (Eq. \eqref{eq:IIA_loss});\\
		\hphantom{~~~~~~} 
		Updating the attribute branch (Eq. \eqref{eq:att_loss_IIA});\\
		\textbf{end for} \\%[0.05cm]
		\textbf{Step II: Unsupervised Target Domain Adaptation (Sec. \ref{sec:method_adaptation})}
		\textbf{for} $t=1$ \textbf{to}  \textsl{max-iteration} \textbf{do} \\ 
		\hphantom{~~~~~~} 
		Sampling a batch of unlabelled target training data; \\
		\hphantom{~~~~~~} 
		Attribute branch evaluation to obtain the soft labels;\\
		\hphantom{~~~~~~} 
		Updating the IIA encoder-decoder (Eq. \eqref{eq:IIA_loss});\\
		\hphantom{~~~~~~} 
		Updating the attribute branch (Eq. \eqref{eq:att_loss_IIA}).\\
		\textbf{end for} %\\
	}
\end{algorithm}

%-------------------------------------------------------------------------

\section{Experiments}

\begin{figure}%[th!]
	\centering
	\includegraphics[width=1\linewidth]{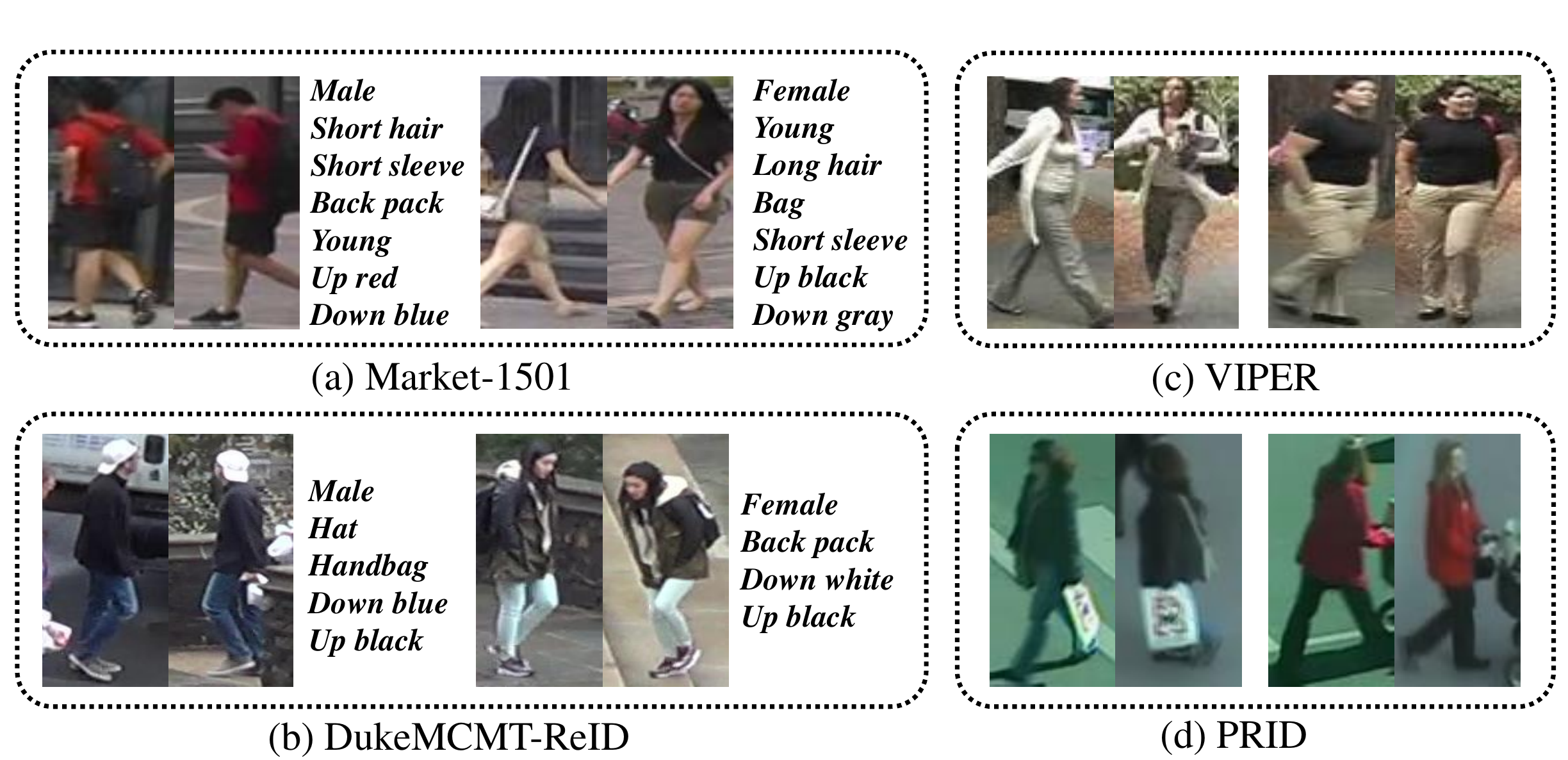}
	\vskip -0.1cm
	\caption{%\footnotesize
	Example of person images and attribute labels.
	Each pair represents two images of the same person.}
	\label{fig:dataset}
%	\vspace{-.2cm}
\end{figure}

\noindent {\bf Datasets and Evaluation Protocol}
We choose four widely adopted person re-id benchmarks for
experimental evaluations  (Fig. \ref{fig:dataset}).
We adopt the standard supervised re-id data split settings 
and only use the test data for model evaluation whilst the training part is ignored.\\
(1) The \textbf{\em Market-1501} dataset \cite{zheng2015scalable} contains 32,668
images of 1,501 pedestrians, each of which was captured
by at most six cameras at a university campus. 
All of the images were cropped by a pedestrian detector
and therefore presenting more challenges to re-id models
due to more background clutters and the misalignment problem.
{\em Evaluation Protocol}:
We used the standard training/test split (750/751) 
and evaluated on single-query evaluation settings
\cite{zheng2015scalable}.\\
(2) The \textbf{\em DukeMTMC-ReID} dataset \cite{zheng2017unlabeled} contains
$2\!\sim\!426$ images per person captured by 8 non-overlapping camera views.
This dataset was constructed from the multi-camera tracking dataset 
DukeMTMC by random selection of manually labelled tracklet bounding boxes. 
{\em Evaluation Protocol}:
We followed \cite{zheng2017unlabeled} by splitting all 1,404 person identities into two halves 702/702 for model training and test, respectively and testing re-id tasks in the single-query setting.
\\
%The raw surveillance
%video data were captured on a university campus.
%
%For target datasets, we add two most representative small scale dataset which has been widely explored by unsupervised re-id methods: VIPeR and PRID:
%
(3) The \textbf{\em VIPeR} dataset \cite{ELF_ECCV08} has 632 identities each with two images 
captured from two camera views in different scenarios of illumination, postures and viewpoints.
This dataset is also featured with low resolution therefore giving rise to an extremely 
challenging re-id task. 
{\em Evaluation Protocol}:
We randomly split the whole population into two halves as training/test sets.
We repeat 10 times of random split and report the average result.\\
%We randomly selecting 10 test sets, and each contains 316 persons.
%
(4) The \textbf{\em PRID} dataset \cite{hirzer2011person} 
consists of person images from two camera views: View
A captures 385 people, whilst View B contains 749 people.
Only 200 people appear in both views.
{\em Evaluation Protocol}:
We use the single shot version
in our experiments
as \cite{zhang2016learning}.
In each data split, 100 people with one image from each view
are randomly chosen from the 200 present in both camera
views as the training set, while the remaining 100 of View
A are used as the probe set, and the remaining 649 of View
B are used as gallery. Experiments are repeated over 10
random splits.

For performance metric, we use the cumulative matching characteristic (CMC) and mean Average Precision (mAP).

\vspace{0.1cm}
\noindent {\bf Attribute Annotation}
In our evaluations, we use either {\em Market-1501} \cite{zheng2015scalable} or
{\em DukeMTMC-ReID)} dataset \cite{lin2017improving} 
as the source domain, since they provide both identity and attribute labels
(Fig. \ref{fig:dataset}).
Specifically, there are 27/23 classes of attributes labelled for Market-1501 / DukeMTMC-ReID \cite{lin2017improving}.
To ensure the unsupervised re-id property, we do not test the Market-1501 when
it is used as the source domain. 
This similarly applies to DukeMTMC-ReID.

%, including: gender
%(male, female), hair length (long, short), sleeve length
%(long, short),length of lower-body clothing (long, short),
%type of lower-body clothing (pants, dress), wearing hat (yes,
%no), carrying bag (yes, no), carrying backpack (yes, no),
%carrying handbag (yes, no), 8 colors of upper-body clothing
%(black, white, red, purple, yellow, gray, blue, green), 9
%colors of lower-body clothing (black, white, pink, purple,
%yellow, gray, blue, green, brown) and age (child, teenager,
%adult, old).
%%
%(2) The \textbf{Duke(DukeMTMC-reID)} dataset \cite{lin2017improving} contains
%2∼426 images per person captured by 8 camera views.
%This dataset was constructed from the multi-camera tracking dataset DukeMTMC by random selection of manually
%labelled tracklet bounding boxes. The raw surveillance
%video data were captured on a university campus.\cite{lin2017improving} annotate 23 attributes, including:
%gender (male, female), shoe type (boots, other shoes), wearing
%hat (yes, no), carrying bag (yes, no), carrying backpack
%(yes, no), carrying handbag (yes, no), color of shoes (dark,
%light), length of upper-body clothing (long, short), 8 colors
%of upper-body clothing (black, white, red, purple, gray,
%blue, green, brown) and 7 colors of lower-body clothing
%(black, white, red, gray, blue, green, brown).

\vspace{0.1cm}
\noindent \textbf{Implementation Details }
We realised the TJ-AIDL model in the Tensorflow framework \cite{abadi2016tensorflow}.
The IIA encoder is designed as a 3-FC-layers network with
their output dimensions as $512$/$128$/$m$ %$D_{1024}-D_{512}-D_{128}-D_{n}$ 
(${m}$ is the number of attribute labels).
A network of a mirror structure is used in the IIA decoder.
We fixed both $\lambda_{1}$ and $\lambda_{2}$ to 10 (Eq. \eqref{eq:IIA_loss})
by scale alignment.
%, to keep the the consist loss and reconstruct loss magnitude comparable with attribute classification loss.
We pre-trained the MobileNet on ImageNet for both identity and attribute branches. 
We used the Adam optimiser~\cite{kingma2014adam} 
with a learning rate of 0.002 and the default momentum terms $\beta_{1}=0.5$, $\beta_{2}=0.999$. 
We set the mini-batch size to 8. 
%\textcolor{red}{
	We started with training the identity branch by 100,000 iterations 
	on the source identity labels and 
	then the whole model by 20,000 iterations for both 
	transferable joint learning on the labelled source data and 
	unsupervised domain adaptation on the unlabelled target data.
        %}
%The mini-batch size is set to 50. 

% and vice versa for the decoder network. 
%
%%learning rate as 0.002,
%We trained the
%models until convergence by setting the maximal iterations 100, 000 for all the momentum term $β1$ = 0.5,
%$β2$ = 0.999. 
%
%For the intermediate and attribute branch,similar setting was applied, the maximal iterations is  20, 000. We set empirically the learning rate as 0.002 with Adam Optimizer for training and 0.0002 for cross domain adaptation.
%
%$w_{1}$ and $w_{2}$ was set to 10, to keep the the consist loss and reconstruct loss magnitude comparable with attribute classification loss.

\begin{table} %[h]
	\footnotesize
	\centering
	\caption{
		Unsupervised re-id performance evaluation. % with state-of-the-art methods.
		{\bf Metric}: Rank-1 and mAP (\%). 
		The $1^\text{st}/2^\text{nd}$ best results are in {\color{red}red} and {\color{blue}blue}.
		TJ-AIDL$^\text{Duke}$ / TJ-AIDL$^\text{Market}$: 
		Our TJ-AIDL using DukeMCMT-ReID and Market-1501 as the labelled source, respectively.
		%domain with labelled identity and attributes.
	}
	\setlength{\tabcolsep}{0.15cm}
	\begin{tabular}{l||c|c|c|c|c|c}
		\hline
		Dataset & VIPeR & PRID & \multicolumn{2}{c|}{Market-1501} & \multicolumn{2}{c}{DukeMCMT} \\
		\hline
		Metric (\%) & R1 & R1 & R1 & mAP & R1 & mAP \\
		\hline
		SDALF\cite{farenzena2010person}   & 19.9 & 16.3 & - &-& - &- \\
		DLLR \cite{kodirov2015dictionary}  & 29.6 & 21.1&- &-& - &-\\		
		CPS \cite{cheng2011custom}  & 22.0 & - &- &-& - &-\\		
		GL  \cite{kodirov2016person} & 33.5& 25.0&- &-& - &- \\
		GTS \cite{wang2014unsupervised}  & 25.2 & - &- &-& - &-\\
		SDC\cite{zhao2017person}   & 25.8 & -&- &-& - &- \\		
		ISR  \cite{lisanti2015person} & 27.0 & 17.0 &40.3 & 14.3& -& - \\
		\hline		
		Dic\cite{BMVC2015_44}  & 29.9 & - & 50.2& 22.7 &- & -\\
		RKSL\cite{wang2016towards}  & 25.8 & - & 34.0& 11.0 & - &-\\
		SAE\cite{lee2008sparse}  & 20.7 & - & 42.4 & 16.2 & - & -\\	
		AML\cite{qin2015unsupervised}  & 23.1 & - & 44.7& 18.4 & - &-\\	
		UsNCA \cite{qin2015unsupervised}  & 24.3 & - &45.2 & 18.9 & - & -\\
		CAMEL \cite{yu2017cross}  & 30.9 & - & \bf\color{blue} 54.5 & \bf \color{blue}26.3 &- & -\\
		PUL \cite{fan2017unsupervised}  & -  & -  & 44.7 & 20.1 & \bf\color{blue} 30.4 & \bf\color{blue}16.4 \\
		\hline		
		kLFDA\_N \cite{xiong2014person}   & 15.9 & 9.1 & - &-&- &-\\
		SADA+kLFDA  \cite{xiong2014person} &15.2& 8.7 & - &- &- &-\\
		AdaRSVM  \cite{ma2015cross} & 10.9& 4.9 &- &- & - &-\\
		UDML  \cite{peng2016unsupervised} & 31.5 & 24.2 &- &- & - &-\\
		SSDAL \cite{su2016deep}  & \bf \color{blue} 37.9 &  20.1& 39.4&19.6 &- &- \\	
		
		\hline
		\bf TJ-AIDL$^\text{Duke}$  
		&35.1  &\bf\color{red}34.8 &\bf\color{red}58.2&\bf \color{red}26.5 & N/A & N/A \\
		\bf TJ-AIDL$^\text{Market}$ 
		& \bf\color{red}38.5 & \bf\color{blue}26.8 & N/A & N/A &\bf\color{red}44.3 &\bf \color{red}23.0 \\
		\hline
	\end{tabular}%
	\label{tab:dataset_art}%
\end{table}%

\subsection{Comparisons to the State-Of-The-Arts}
We compare 19 models in three categories of existing unsupervised re-id methods:
{\bf (1)} Hand-crafted feature based methods {\em without} transfer learning: 
SDALF\cite{farenzena2010person}  and CPS \cite{cheng2011custom}, those
features are designed to be view invariant. Dictionary Learning based
methods DLLR \cite{kodirov2015dictionary}, graph-learning-based model
GL \cite{kodirov2016person}, sparse representation learning methods
ISR  \cite{lisanti2015person}, salience-learning-based GTS
\cite{wang2014unsupervised} and SDC\cite{zhao2017person}. 
{\bf (2)} 
Source identity knowledge transfer learning based methods:
Dic \cite{BMVC2015_44},
RKSL \cite{wang2016towards},
SAE \cite{lee2008sparse},  
AML \cite{qin2015unsupervised},
UsNCA \cite{qin2015unsupervised},
CAMEL \cite{yu2017cross}.
{\bf (3)} Source identity and attribute knowledge based transfer methods:
kLFDA\_N \cite{xiong2014person} 
SADA+kLFDA \cite{xiong2014person}
AdaRSVM  \cite{ma2015cross} 
UDML \cite{peng2016unsupervised}.

Table \ref{tab:dataset_art} shows that:
{\bf(1)} Our method outperforms clearly 
all existing state-of-the-art models, % in most cases,
improving the Rank-1 by 
0.6\% (38.5-37.9), 9.8\% (34.8-25.0), 
3.7\% (58.2-54.5), 13.9\% (44.3-30.4)
over the best alternative method 
on VIPeR/PRID/Market-1501/DukeMCMT-ReID, respectively.
This suggests the overall performance advantages of the proposed
TJ-AIDL in the capability of multi-source (attribute and identity) information extraction
and fusion for cross-domain unsupervised re-id matching.
{\bf(2)} When compared to the existing methods of $1^\text{st}$ category (non-learning based) % (i.e. first category),
the performance margins are even much larger, 
e.g. the Rank-1 boost is
8.9\% (38.5-29.6),
9.8\% (34.8-25.0),
17.9\% (58.2-40.3)
on VIPeR/PRID/Market-1501, respectively.
This indicates the importance of learning from labelled source supervision
in cross-domain re-id scenarios, since hand-crafted features are
not sufficiently generalisable across different domains with varying camera view conditions.
{\bf(3)} When comparing the methods between $2^\text{nd}$ (identity transfer) and 
$3^\text{rd}$ (identity and attribute joint transfer) category,
it is interestingly found that the latter is not necessarily
superior over the former.
This means that using more supervision in cross-domain transfer learning
is non-trivial particularly when the label property is heterogeneous such as
identity and attribute.
This also indirectly suggest the model design advantages of our TJ-AIDL
in exploiting the diverse knowledge in different types of label data
for the more challenging cross-domain re-id tasks in 
the unlabelled target scenario typical in real-world deployments.

Finally, it is worth noting that
the performance advantages by our TJ-AIDL are achieved using
much less supervision data of lower diversity from only {\em one} source domain
(16,522 images of 702 identities/classes on DukeMCMT-ReID, or
12,936 images of 751 identities on Market-1501)
than strong existing competitors. 
For example,
the methods of $2^\text{nd}$ category utilise 
$7$ different person re-id datasets with high domain varieties 
(CUHK03\cite{li2014deepreid}, CUHK01\cite{li2012human}, PRID, VIPeR, 3DPeS\cite{baltieri20113dpes}, i-LIDS\cite{prosser2010person}, Shinpuhkan\cite{kawanishi2014shinpuhkan2014}) 
including a total of 44,685 images and 3,791 identities;
The UDML \cite{peng2016unsupervised} 
exploits three different source domains  % among VIReR, PRID, CUHK01 and Market-1501
including 46,966 images of 3,246 identities for test on VIPeR (target),
and 
47,096 images of 3,493 identities for test on PRID (target).
% three of four dataset like for viper test:385+1360+1501=3,246 ID
%SSDAL \cite{su2016deep}transfer from PETA source, which include 10 diverse datasets annotate with 8705 person (19,000 images), and MOT dataset with 20,000  images with ID.
% VIPER 1264  Prid 385+749 cuhk03 13164 market 32668
% for PRID: source 3493 id   47096 image
% for VIPer: source 3246 ID 46966 images
%
%
The SSDAL \cite{su2016deep} 
% transfer from PETA source, which include 
benefits from $10$ diverse datasets consisting in 
19,000 images of 8,705 person identities and another
20,000 images of 1,221 person tracklets. % tracking. 

\subsection{Comparisons to Alternative Fusion Methods}
\label{sec:exp_fusion_method}

\begin{table*} [htbp]
%	\footnotesize
	\centering
	\caption{
		Comparing different multi-source fusion methods.
%		(Market-to-Duke/ Duke-to-Market)
	}
\setlength{\tabcolsep}{0.15cm}
	\begin{tabular}{c||cccc|c||cccc|c}
		\hline
		 Source $\rightarrow$ Target     
		 &\multicolumn{5}{|c||}{Market-1501 $\rightarrow$ DukeMCMT-ReID} 
		 &\multicolumn{5}{|c}{DukeMCMT-ReID $\rightarrow$ Market-1501} \\ \hline
		Metric (\%) & Rank1 & Rank5 & Rank10 & Rank20 & mAP & Rank1 & Rank5 & Rank10 & Rank20 & mAP \\ 
		\hline		
		Independent Supervision    
		&  33.8 & 49.5 & 56.0 &63.8 & 16.9 
		& 54.9 &72.9 &79.3 &85.2 & 24.5   \\
%		CCA    &   &   &   &   
%		&    &   &    &      \\
%		GTNN\cite{hu2017attribute}    &   &   &   &   
%		&   &  &   &    \\
		%joint\_train 
		Joint Supervision
		& 37.9 &  52.1 &   58.6 &65.3 &  20.6
		 &  53.4   &71.2 &  78.1 &83.3 &  21.9  \\
		 \hline
		\bf TJ-AIDL & \textbf{44.3} & \textbf{59.6} & \textbf{65.0} &\textbf{70.0}  & \textbf{23.0} 
		& \textbf{58.2}  &  \textbf{74.8} &  \textbf{81.1} &\textbf{86.5} & \textbf{26.5} \\	
		\hline
	\end{tabular}%
	\label{tab:fusion}%
\end{table*}%

We compare the TJ-AIDL with two multi-source fusion methods:
{\bf (a)} 
{\footnotesize \textsf{Independent Supervision}}: 
Independently train a
deep CNN model for either attribute or identity label 
in the source domain % (Figure \ref{fig:baseline} (a)); 
and
use the concatenated feature vectors of the two models
for re-id matching in the target domain.
{\bf (b)} {\footnotesize \textsf{Joint Supervision}}: 
A seminal multi-task joint learning CNN framework
subjecting the identity and attribute supervision to
a shared feature representation 
%the fusion of all the scale-specific features 
in the end-to-end model training.
%(Figure \ref{fig:baseline} (b)).
For re-id deployment on the target domain, we use the multi-supervision shared feature representation.

Table \ref{tab:fusion} shows that: 
{(1)} 
The TJ-AIDL outperforms both alternative fusion methods.
This suggests a clear advantage of our method %multi-scale consensus learning and regularisation
in exploiting and fusing multiple supervision for 
cross-domain re-id in an unsupervised manner.
{(2)} 
Our method achieves more performance gain over the competitors 
on the transfer from Market-1501 (source) to DukeMTMC-ReID (target)
than the opposite transfer.
This is expected and reasonable because
relative to Market-1501, person images from DukeMTMC-ReID 
have more changes in image resolution and background clutter due to 
wider camera views and more complex scene layout, 
which means the source information itself from Market-1501
is insufficient to generalise the target DukeMTMC-ReID setting
and therefore leading to a higher need for domain adaptation.
Our model strongly and naturally meets this deployment requirement.
For the opposite transfer from DukeMTMC-ReID to Market-1501,
our model gives less performance gain
since there is a lower need for domain adaptation.

\subsection{Further Analysis and Discussions}

\begin{table} [htbp]
%	\footnotesize
	\centering
	\caption{
		Complementary of identity-discriminative and attribute-sensitive features
		learned by the proposed TJ-AIDL.
		}
		\setlength{\tabcolsep}{0.05cm}
	\begin{tabular}{c||cccc|c}
		\hline
		 Source $\rightarrow$ Target       
		 &\multicolumn{5}{|c}{Market-1501 $\rightarrow$ DukeMCMT-ReID} 
		 \\ \hline
		Metric (\%) & Rank1 & Rank5 & Rank10 & Rank20 & mAP \\
		\hline		
		Attribute Only  & 24.3 & 38.3 &  45.7 &53.0 & 10.0
		 \\
		ID Only  & 30.6 & 44.9 &  50.5 & 59.3& 14.6
		 \\
		Attribute + ID ({\bf Full}) & \textbf{44.3} & \textbf{59.6} & \textbf{65.0}&\textbf{70.0} & \textbf{23.0 } 
		\\\hline \hline
		Source $\rightarrow$ Target & \multicolumn{5}{|c}{DukeMCMT-ReID $\rightarrow$ Market-1501} 
		\\ \hline
		Attribute Only
		& 38.0  & 59.2  & 67.6 & 75.7& 13.6 \\
		ID Only
		&51.6 &69.8 & 76.6 &81.6 & 21.6  \\
		Attribute + ID ({\bf Full})
		& \textbf{58.2}  &  \textbf{74.8} &  \textbf{81.1} &\textbf{86.5} & \textbf{26.5} \\
		\hline
	\end{tabular}%
	\label{tab:source}%
\end{table}%

\begin{figure}%[th!]
	\centering
	\includegraphics[width=1\linewidth]{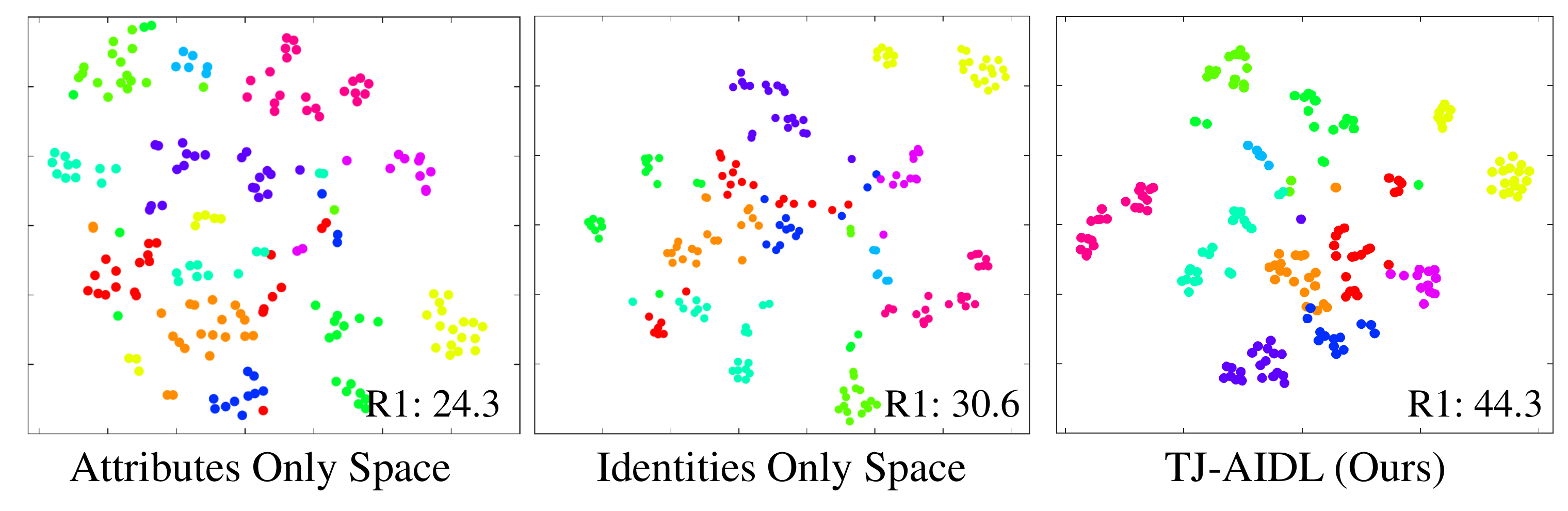}
	%\vskip -0.5cm
	\caption{%\footnotesize
	 Feature distributions of 10 random test identities in three transferred feature spaces (Market-1501 $\rightarrow$ DukeMCMT-ReID) visualised by t-SNE \cite{maaten2008visualizing}. 
%	 Different identities are shown in different colours. 
	 Colour coded identity classes. 
%	 Best viewed in colour.
 }
	\label{fig:tsne}
	%\vspace{-.1cm}
\end{figure}

\noindent {\bf Effect of Joint Attribute and Identity Features}
We evaluated the effect of joint attribute and identity features by
comparing their individual re-id performances against that of the joint feature.
We obtain their individual model by training a MobileCNN
using either identity or attribute label only.
Table \ref{tab:source} shows 
feature representation learned by only one supervision is 
significantly inferior that that by our TJ-AIDL. 
For instance, the TJ-AIDL feature 
outperforms ID Only 
by $13.7\%$(44.3-30.6) in Rank-1 
and $8.4\%$ (23.0-14.6) in mAP on DukeMCMT-ReID (target);
by $6.6\%$ (58.2-51.6) in Rank-1
and $4.9\%$(26.5-21.6) in mAP on Market-1501 (target).
These validate the complementary effect of jointly learning
attribute and identity information 
and importantly strong capability of our model in maximising 
this latent information in a more transferable context. We also plot three feature distributions of 10 randomly selected test identities of DukeMTMC-ReID (transferred from Market-1501). Figure \ref{fig:tsne} shows that: (1) Neither transferring the knowledge of attributes or identities alone can form per-identity compact clusters; (2) By our TJ-AID that transfers attributes and identities jointly, the feature distributions of 10 test identities are much more separated.

%by our IIA model. 

%In table.\ref{tab:adaptation}, we observe that our collaborative learning based adaptation methods performs well on both Market and Duke dataset, moreover, our model has superior performance on more diverse and changeable environment considering Duke dataset has more diverse scale problem.

%\vspace{-.1cm}

\begin{table} [htbp]
%	\footnotesize
	\centering
	\caption{
		Effect of the target domain adaptation in TJ-AIDL.}
\setlength{\tabcolsep}{0.08cm}
	\begin{tabular}{c||cccc|c}
		\hline
		Source $\rightarrow$ Target & \multicolumn{5}{|c}{Market-1501 $\rightarrow$ DukeMCMT-ReID} 
		\\ \hline
		Metric (\%) & Rank1 & Rank5 & Rank10 & Rank20 & mAP \\
		\hline		
		%ours\_A(no\_ad)& 35.5 &  50.9 &  56.6 & 17.6
		%  & 49.6    & 68.5 & 76.2 &20.9 \\
		{\bf w/o} Adaptation  
		& 39.6  &  55.5 &  62.2 &67.5 & 22.0
		 \\
		{\bf w} Adaptation & \textbf{44.3} & \textbf{59.6} & \textbf{65.0} & \textbf{70.0} & \textbf{23.0} 
		\\ \hline \hline
		Source $\rightarrow$ Target & \multicolumn{5}{|c}{DukeMCMT-ReID $\rightarrow$ Market-1501} 
		\\ \hline
		{\bf w/o} Adaptation 
		& 57.1 & 74.4 & 80.4 &85.7  & 26.2 
		\\ \hline
		{\bf w}  Adaptation 
		& \textbf{58.2}  &  \textbf{74.8} &  \textbf{81.1} &\textbf{86.5} & \textbf{26.5} \\
		\hline
	\end{tabular}%
	\label{tab:adaptation}%
\end{table}%

\vspace{0.02cm}
\noindent {\bf Effect of Target Domain Adaptation}
We evaluated the effect of 
the attribute consistency driven 
domain adaptation on unlabelled target training data. 
Table \ref{tab:adaptation} shows that 
this adaptation clearly improves the re-id performance for the transfer
of DukeMCMT-ReID $\rightarrow$ Market-1501
($1.1\%$ Rank-1 boost)
and more significantly for 
the case of Market-1501 $\rightarrow$ DukeMCMT-ReID
($4.7\%$ Rank-1 boost).
This shares a similar observation and underlying reason as in Table \ref{tab:fusion},
validating the benefit of our method 
in varying cross-domain model adaptation in
improving the model compatibility when deployed
to a new target scenario.

%\begin{table} [htbp]
%%	\footnotesize
%	\centering
%	\caption{
%		Effect of the Reconstruction Loss (RL) in TJ-AIDL.}
%	\setlength{\tabcolsep}{0.1cm}
%	\begin{tabular}{c||cccc|c}
%		\hline
%		Source $\rightarrow$ Target & \multicolumn{5}{|c}{Market-1501 $\rightarrow$ DukeMCMT-ReID} \\
%		\hline
%%		 Dataset     &\multicolumn{4}{|c||}{Market-to-Duke} &\multicolumn{4}{|c}{Duke-to-Market} \\ \hline
%		Metric (\%) & Rank1 & Rank5 & Rank10 & Rank20 & mAP \\
%		\hline		
%		%ours\_A(no\_ad)& 35.5 &  50.9 &  56.6 & 17.6
%		%  & 49.6    & 68.5 & 76.2 &20.9 \\
%		{\bf W/O} RL
%		&    &   &    &  &  
%		\\
%		{\bf W} RL
%		& \textbf{44.3} & \textbf{59.6} & \textbf{65.0} & & \textbf{23.0 } 
%		\\
%		\hline \hline
%		Source $\rightarrow$ Target & \multicolumn{5}{|c}{DukeMCMT-ReID $\rightarrow$ Market-1501} 
%		\\ \hline
%		{\bf W/O} RL
%		& 56.5 & 73.2 & 79.6  & 85.7 & 25.0  \\
%		{\bf W} RL
%		& \textbf{58.2} &  \textbf{74.8} &  \textbf{ 81.1 } & & \textbf{26.5 } \\
%		\hline
%	\end{tabular}%
%	\label{tab:dataset_compare}%
%\end{table}%

%\vspace{0.02cm}
%\noindent {\bf Effect of Target Domain Adaptation}
%We finally evaluated the bene

%\subsection{Cross Dataset Person Re-ID with Domain Adaptation}
%------------------------------------------------------------------------

\section{Conclusion}

We presented a novel Transferable Joint Attribute-Identity Deep Learning (TJ-AIDL)
for more discriminative joint learning of the identity and attribute supervision from an auxiliary domain
%in an end-to-end manner
in order to particularly address the scalable unsupervised person re-identification problem
in the context of heterogeneous multi-task joint learning and domain
transfer learning.
In contrast to most existing re-id methods
that either ignore the scalability issue in re-id or 
exploit a straightforward yet sub-optimal multi-task joint learning of multi-supervision, 
the proposed model is capable of transferring and integrating multiple
heterogeneous supervision and maximising their latent compatibility 
for optimal person re-id in a progressive and more transferable means.
This is achieved by introducing an Identity Inferred Attribute space 
for interactive attribute and identity discriminative learning in a two-branches CNN architecture.
Moreover, we introduce an attribute consistency maximisation mechanism
to further discriminatively adapt a learned TJ-AIDL model 
to fit any given target re-id deployment without the need for additional data labelling
and hence very scalable to real-world applications.
Extensive evaluations were conducted on four re-id benchmarks to validate the advantages of
the proposed TJ-AIDL model over a wide range of state-of-the-art methods 
on different re-id task scenarios with various challenges.
We also 
compared the TJ-AIDL model with popular multi-supervision fusion methods and
provided detailed component analysis with insights into 
the performance gain of our model design.

\section*{Acknowledgements}
%\vspace{-0.2cm}
\noindent This work was partially supported by the China Scholarship Council, 
Vision Semantics Ltd, Royal Society Newton Advanced Fellowship Programme (NA150459),
and InnovateUK Industrial Challenge Project on Developing and Commercialising Intelligent Video Analytics Solutions for Public Safety.

{\small
\bibliographystyle{ieee}
\bibliography{transfer}
}

\end{document}